\documentclass[letterpaper, 10 pt, conference]{ieeeconf}  % Comment this line out if you need a4paper

\IEEEoverridecommandlockouts                              % This command is only needed if 
\title{\LARGE \bf
The Value of Planning for Infinite-Horizon Model Predictive Control}

\author{Nathan Hatch$^*$ and Byron Boots$^*$% <-this % stops a space
\thanks{$^*$University of Washington, Seattle WA 98105 USA.
        {\tt\small \{nhatch2, bboots\}@cs.washington.edu}
}}

\usepackage{cite}
\usepackage{amsmath,amssymb,amsfonts}
\usepackage{algorithmic}
\usepackage{graphicx}
\usepackage{textcomp}
\usepackage{xcolor}
\usepackage{comment} 
\usepackage{makecell}
\usepackage[colorlinks=true,linkcolor=blue]{hyperref}
\def\BibTeX{{\rm B\kern-.05em{\sc i\kern-.025em b}\kern-.08em
    T\kern-.1667em\lower.7ex\hbox{E}\kern-.125emX}}

\newcommand{\matx}[2]{\left[ \begin{array}{#1} #2 \end{array} \right]}

\DeclareMathOperator*{\argmin}{arg\,min}

\newcommand{\argminprob}[1]{\underset{#1}{\argmin}}

\newcommand\blfootnote[1]{%
  \begingroup
  \renewcommand\thefootnote{}\footnote{#1}%
  \addtocounter{footnote}{-1}%
  \endgroup
}

\begin{document}

\maketitle
\thispagestyle{empty}
\pagestyle{empty}

\begin{abstract}
Model Predictive Control (MPC) is a classic tool for optimal control of complex, real-world systems.
Although it has been successfully applied to a wide range of challenging tasks in robotics,
it is fundamentally limited by the prediction horizon, which, if too short, will result in myopic decisions.
Recently, several papers have suggested using a learned value function as the terminal cost for MPC.
If the value function is accurate, it effectively allows MPC to reason over an \emph{infinite} horizon.
Unfortunately, Reinforcement Learning (RL) solutions to value function approximation can be difficult to realize for robotics tasks.
In this paper, we suggest a more efficient method for value function approximation that applies to goal-directed problems, like reaching and navigation.
In these problems, MPC is often formulated to track a path or trajectory returned by a planner.
However, this strategy is brittle in that unexpected perturbations to the robot will require replanning, which can be costly at runtime.
Instead, we show how the intermediate data structures used by modern planners can be interpreted as an approximate \emph{value function}.
We show that that this value function can be used by MPC \emph{directly}, resulting in more efficient and resilient behavior at runtime.\blfootnote{\copyright 2021 IEEE. Personal use of this material is permitted. Permission from IEEE must be obtained for all other uses, in any current or future media, including reprinting/republishing this material for advertising or promotional purposes, creating new collective works, for resale or redistribution to servers or lists, or reuse of any copyrighted component of this work in other works.}
\end{abstract}

\section{INTRODUCTION}
\label{sec:introduction}
In this paper, we revisit the classic planning and control paradigm ubiquitous in robotics.
In this paradigm, behavior is generated \emph{hierarchically}. First, a planner computes a high-level solution to a simplified global problem, typically solved offline and in advance. Then, at runtime, a controller follows that plan locally using a more detailed dynamics model. This strategy arises from the difficulty of solving the entire behavior generation approach at once. Often, this approach works very well \cite{ferguson2008motion,kuwata2009real}.

In some applications, this approach can be brittle. Global information is often condensed into a \emph{single} plan and the controller's cost is defined relative to that information.
Stochasticity in the environment, or mismatch between the models used for planning and control, can mean that the controller deviates irrecoverably from the plan.
When possible, it is common to rerun the planner from the current location. This can be dangerous or expensive if replanning takes too long.

Compare the above approach to reinforcement learning (RL)~\cite{sutton2018reinforcement}. RL typically attempts to tackle the entire problem at once, building a \emph{global} value function or policy that encodes the best action at any state in the environment. Compared with plans, value functions are much more informative.
A controller with access to a value function could resolve the best immediate action based on value \emph{anywhere} in the state space, resulting in more robust behavior and eliminating the need to replan. While some robotics problems have been successfully solved with RL~\cite{akkaya2019solving,riedmiller2007learning,heess2017emergence}, learning an accurate value function over the entire space space is typically infeasible due to computational and modeling limitations.

We would like to find an approach that lies between these extremes:
more robust than following a single plan, but less expensive than computing a full value function. Recent work on combining Model Predictive Control (MPC) with RL has indeed started to move in this direction. 
MPC is a classic tool for controlling complex, real-world systems that solves an online optimization problem to choose an action that will minimize predicted future cost. While MPC has been applied to a wide range of challenging tasks in robotics~\cite{abbeel2010autonomous,williams2016aggressive,erez2013integrated}, it is fundamentally limited by the prediction horizon, which, if too short, will result in myopic decisions.

Recently, several papers have suggested using a learned value function as the terminal cost for MPC~\cite{chen1998quasi,erez2012infinite,de2008waypoint}. In particular, \cite{erez2012infinite} calculates offline the exact value function along a limit cycle,
then uses this to compute a piecewise quadratic approximation to the value function everywhere. Since the resurgence of deep reinforcement learning, other authors have investigated approximating value functions for MPC using neural networks. For example, POLO \cite{lowrey2018plan} uses a kind of fitted value iteration where the regression targets are found using trajectory optimization.
Another recent example is model-predictive Q-learning (MPQ) \cite{bhardwaj2020information}, which shows a theoretical connection between soft Q-learning \cite{haarnoja2017reinforcement} and information-theoretic MPC \cite{williams2017information}. Like POLO, MPQ learns a value function with the help of trajectory optimization. Other examples include~\cite{anthony2017thinking,silver2016mastering,silver2017mastering,sun2018truncated}.
All of these approaches incorporate global value function information into MPC, effectively allowing MPC to reason over an \emph{infinite} horizon.

Unfortunately, RL solutions to value function approximation can be infeasible for many robotics tasks.
For goal-directed navigation or reaching tasks, using approaches like POLO~\cite{lowrey2018plan} or MPQ~\cite{bhardwaj2020information} to compute a terminal cost for MPC makes little sense. These methods are time consuming and expensive, even for simple problems. By contrast, planning algorithms can generate a rough trajectory quickly. This leads to a natural question: can planning algorithms be exploited to generate better, more robust terminal cost functions for MPC? This leads to the key insight in this work: modern planning algorithms conveniently already compute an approximate value function as an intermediate step toward generating a solution. Consider, for example, RRT* \cite{karaman2010optimal} searching backwards from the goal.
This algorithm produces not just a path from the starting location, but
also an entire tree of paths to the goal,
each node annotated with its distance from the goal---an approximate value function.
The algorithm RRT\# \cite{arslan2013use} makes this connection even more explicit by fully solving the Bellman optimality equations at each iteration, to ensure that the value function is the optimal one for the current planning graph.

In this work, we propose using entire planning trees as approximate value functions that can be used as the terminal cost in MPC. This results in a strategy that is more robust than the classic robotics planning and control paradigm, but much faster than RL approaches to the same problems. Specifically, we focus on the interface between two commonly used algorithms in robotics, the stochastic MPC algorithm Model Predictive Path Integral control (MPPI)~\cite{williams2017information} and the sampling-based planning algorithm RRT\#~\cite{arslan2013use}.
We propose treating the terminal point of each sampled MPC rollout as a node of the planning tree and using the corresponding cost-to-go as the terminal cost in MPC. In other words, we use the value function computed as a by-product of planning to extend the horizon of MPC. This information improves the performance of MPC in several ways.
First, MPC can be made more \emph{accurate}, as each terminal state can be evaluated using the most relevant planning node.
Second, MPC can be made more \emph{robust}.
When the controller is perturbed, it simply uses existing nodes in the planning tree---there is no replanning time.
Third, the value returned by RRT\#, while approximate, lies in a sweet spot: it is fast to compute and useful to MPC in practice.

\section{BACKGROUND}

\subsection{Bellman Backups on Directed Graphs}
\label{sec:bellman}

We begin by providing a brief background discussion on value iteration and its relation to  planning algorithms.
In RL, Bellman backups are used by value iteration (among other algorithms) to compute a value function.
In the simplest setting, value iteration works on a discrete state space $\mathcal{S}$ and action space $\mathcal{A}$.
For each state $s \in \mathcal{S}$, each action $a \in \mathcal{A}$ has some cost $c(s,a)$
and deterministically transitions to a new state $f(s,a)$.
To compute the value function $V:\mathcal{S} \rightarrow \mathbb{R}$, we perform sweeps through the states, computing an (undiscounted) Bellman backup at each state:
\begin{align}
    V(s) \leftarrow \min_a c(s,a) + V(f(s,a))
\end{align}
Equivalently, we can formulate this problem as finding the minimum-cost path along a weighted, directed graph.
The vertices of the graph are the states $\mathcal{S}$, and the edges are the pairs $(s,f(s,a))$ for $a \in \mathcal{A}$,
with edge weight $\hat c(s,f(s,a)) := c(s,a)$.
Let $\mathcal{N}(s)$ denote the out-neighbors of $s$.
Then the Bellman backup looks like this:
\begin{align}
\label{eq:bellman}
    V(s) \leftarrow \min_{s' \in \mathcal{N}(s)} \hat c(s,s') + V(s')
\end{align}

\subsection{Connection between Bellman Backups and Planning}

Finding the minimum-cost path along a directed graph is precisely what planning algorithms do.
However, they vary in how they construct this graph.
In this paper, we focus on the stochastic graphs constructed by RRT* \cite{karaman2011sampling}.
Briefly, RRT* iteratively constructs a graph by sampling states in configuration space and connecting them to all existing vertices within some search radius.
It then performs a Bellman backup for the new vertex and all of the neighboring vertices.

Much prior work has investigated the connection between graph-based planning and Bellman backups.
Like all value iteration methods, RRT* is not guaranteed to produce a globally consistent value function after each Bellman backup.
To address this, RRT\#~\cite{arslan2013use} incorporates graph search techniques from D* \cite{stentz1997optimal} to efficiently update the entire value function after adding each new vertex.
This line of reasoning has produced a number of follow-up works; e.g. \cite{arslan2016incremental,arslan2017sampling}.

In LPA* \cite{koenig2004lifelong},
an incremental heuristic search technique like D*,
Bellman backups are instead called ``vertex expansions''.
A vertex is considered ``consistent'' in LPA* precisely when it satisfies the local Bellman optimality equation---that is,
when performing a Bellman backup does not change the value of the vertex.
After performing a Bellman backup, LPA* decides which neighboring vertices might now be inconsistent,
and adds those vertices to a priority queue.

\section{APPROACH}
\label{sec:approach}
In this work, we consider behavior generation problems in robotics that are typically solved by invoking a planner to generate a path and a closed-loop controller that follows the path online. This includes many \emph{goal-directed} problems including reaching (in manipulation) and vehicle navigation.

We will frame these problems formally as \emph{reinforcement learning}.
Specifically, we consider the problem of learning a policy in an infinite-horizon undiscounted Markov Decision Process (MDP) with terminal states.
Such an MDP is defined by a tuple
$\mathcal{M} = \{ \mathbb{S}, \mathbb{A}, c, f, s_{start}, S_{goal}\}$
where $\mathbb{S}$ is the state space,\footnote{We use $\mathbb{S}$ and $\mathbb{A}$ to denote continuous state and action spaces, as opposed to $\mathcal{S}$ and $\mathcal{A}$ for discrete ones.}
$\mathbb{A}$ is the action space,
$c(s,a)$ is the per-step cost function,
$s_{t+1} \sim f(s_t, a_t)$ is the stochastic transition function, $s_{start}$ is the start state, and $S_{goal} \subseteq \mathbb{S}$ is the goal region.
A policy $\pi(\cdot|s)$ outputs a distribution over actions given a state. Let $\mu^{\pi}_{\mathcal{M}}$ be the distribution over state-action trajectories obtained by running policy $\pi$ on $\mathcal{M}$. 
The value function for a given policy $\pi$ is defined as
\[\textstyle
V^{\pi}_{\mathcal{M}}(s) = \mathbb{E}_{\mu^{\pi}_{\mathcal{M}}} \left[ {\sum_{t=0}^\infty c(s_t, a_t) \mid s_{0}=s} \right]
\]
and the action-value function as
\[\textstyle
Q^{\pi}_{\mathcal{M}}(s,a) = \mathbb{E}_{\mu^{\pi}_{\mathcal{M}}} \left[ {\sum_{t=0}^\infty c(s_t, a_t) \mid s_{0}=s, a_{0}=a} \right]
\]
Although we have written these as infinite undiscounted sums,
the MDP terminates once $s_t \in S_{goal}$ for some $t$,
after which time no cost is accumulated.
The objective is to find an optimal policy $\pi^* = \argminprob{\pi}\; V^{\pi}_{\mathcal{M}}(s_{start})$. 

\subsection{An RL Perspective on Model Predictive Control}
MPC is widely used in robotics and can be viewed as an online learning approach to synthesizing action sequences for MDPs~\cite{Wagener19a}. Instead of solving for a single globally optimal policy that prescribes the action to take at any state, MPC is a pragmatic approach of optimizing simple, local action sequences at test time.  At each timestep, MPC uses an approximate transition model to search for an action sequence that minimizes cost over a finite horizon. The first action from the sequence is executed on the system, and the process is then repeated from the next state.

Following~\cite{anonymous2021blending}, we formalize this process as  solving a surrogate MDP whose parameters may differ from the parameters of the true underlying MDP that we wish to solve.
The surrogate MDP is $\mathcal{\widehat M} = \{ \mathbb{S}, \mathbb{A}, c, \hat f, s_{start}, S_{goal}, H\}$ with approximate or simplified dynamics $\hat f$ and, importantly, a finite horizon $H$. $\mathcal{\widehat M}$ is repeatedly solved at every state encountered, resulting in a receding horizon method.

In many problems, solving $\mathcal{\widehat M}$ with the limiting horizon $H$ would result in myopic behavior with respect to the original MDP $\mathcal{M}$. This can be especially pronounced in tasks with sparse rewards and the goal-directed tasks considered in this paper,
which naturally require reasoning about longer horizons. 
To contend with the limited horizon, \emph{infinite horizon} MPC~\cite{bhardwaj2020information} sets the terminal cost in MPC as a value function $\widehat{V}$ that adds global information to the problem.

MPC can, therefore, be viewed as iteratively constructing an \emph{estimate} of the Q-function of the original MDP $\mathcal{M}$, under the policy $\pi_{\phi}$ induced by the action sequence $\phi$~\cite{anonymous2021blending}:
\begin{small}
\begin{align}\textstyle
	\label{eq:mpc_q_function}
	Q^\phi_H(s,a) = \mathbb{E}_{\mu^{\pi_\phi}_{\mathcal{\widehat M}}}\left[{\sum_{i=0}^{H-1} c(s_i, a_i) +  \widehat V(s_H) \mid s_{0}=s, a_{0}=a}\right] 
\end{align}
\end{small}

\noindent MPC then iteratively optimizes this estimate (at current system state $s_t$) to update the action sequence
\begin{equation}
    \label{eq:mpc_policy_opt}
    \phi^*_t = \argminprob{\phi} \; Q^\phi_H(s_t, \pi_\phi(s_t)) 
\end{equation}
Several popular MPC algorithms, including MPPI~\cite{anonymous2021blending} and receding-horizon linear quadratic regulators~\cite{Wagener19a} can be viewed through this lens.

\subsection{Adding Value from Planning}
The main challenge is obtaining the value function $\widehat V$  in Eq.~\ref{eq:mpc_q_function}. Previous approaches have attempted to learn it from data via Q-learning~\cite{lowrey2018plan,bhardwaj2020information}. This can be costly, as it requires a large number of interactions with the system. For goal-directed problems, it is often significantly cheaper to find a \emph{plan} that MPC can follow to the goal \cite{kuwata2009real,ferguson2008motion}.
As discussed in the introduction, if cost-to-go values along this path are provided by the planner, then it implicitly defines an approximation of the value function along the path.
If the approximation is ``good enough,'' then we can use it as the terminal cost $\widehat V$ in Eq.~\ref{eq:mpc_q_function}. 
Our key insight is that many modern planning algorithms construct a \emph{planning tree} over a much larger subset of the state space as an intermediate step toward finding the optimal plan, thereby improving the value function estimate.
The connection between planning and RL has been made in many prior works, but none of these have discussed the connection with MPC~\cite{koenig2004lifelong,arslan2013use,arslan2016incremental,arslan2017sampling}.

\section{Algorithm}
In our experimental results in Section~\ref{sec:experiments}, we use MPPI, a stochastic MPC algorithm, as our closed-loop controller \cite{williams2017information}. We use RRT\# to compute an approximate value function via a planning tree that grows backwards from the desired configuration to the current robot configuration \cite{arslan2013use}. The implementation details of these algorithms are provided below.
Please refer to Fig. \ref{alg:notation} for notation.

\begin{figure}[t]
\centering
\framebox{\parbox{3in}{
\begin{algorithmic}
\REQUIRE Configuration space $\mathbb{S}$
\REQUIRE Free space $\mathbb{S}_{free} \subseteq \mathbb{S}$
\REQUIRE Control space $\mathbb{A}$
\REQUIRE Horizon $H$
\REQUIRE Stochastic dynamics $f: \mathbb{S} \times \mathbb{A} \rightarrow \mathbb{S}$
\REQUIRE Model $\hat f: \mathbb{S} \times \mathbb{A}^H \rightarrow \mathbb{S}^{H+1}$
\REQUIRE True cost $c:\mathbb{S} \times \mathbb{A} \rightarrow [0,\infty)$
\REQUIRE Kinematic planning cost $\hat c:\mathbb{S} \times \mathbb{S} \rightarrow [0,\infty)$
\REQUIRE Maximum steering radius $M$
\REQUIRE Maximum search radius $R > M$
\REQUIRE Subroutine $\texttt{nearest}(s,\mathcal{S},\hat c,L,r)$: Returns the $L$ points in the set $\mathcal{S}$ that are nearest to $s$ according to the cost function $\hat c$. Excludes any points $s'$ for which $\hat c(s,s') > r$.
\end{algorithmic}}}
\caption{Notation used in algorithms.}\vspace{-4mm}
\label{alg:notation}
\end{figure}

\subsection{RRT\#}
The RRT\# planning algorithm as used in this paper is shown in Fig. \ref{alg:rrt}.
The pseudocode requires two additional subroutines:
\begin{itemize}
    \item $\texttt{steer}(s_{sample}, s_{near}, \hat c, M)$: Projects the point $s_{sample}$ to the ball of radius $M$ around $s_{near}$ according to the distance function $\hat c$.
    \item $\texttt{value\_iterate}(\mathcal{S},\mathcal{E},\mathcal{V},\hat c)$: Performs sweeps of Bellman backups (Eq. \ref{eq:bellman}) on the graph $(\mathcal{S},\mathcal{E})$, with edge weights defined by $\hat c$.
    Returns the optimal value function.
    The parameter $\mathcal{V}$ is used to initialize the value function for faster convergence.
\end{itemize}

\begin{figure}[t]
\centering
\framebox{\parbox{3in}{
\begin{algorithmic}
\REQUIRE Start and goal states $s_{start}, s_{goal} \in \mathbb{S}$
\REQUIRE Distribution $P$ over $\mathbb{S}$ such that $P(s_{start}) > 0$
\STATE Initialize vertices $\mathcal{S} = \{s_{goal}\}$
\STATE Initialize edges $\mathcal{E} = \{\}$
\STATE Initialize value function $\mathcal{V} = \{(s_{goal},0)\}$
\WHILE {$s_{start} \notin \mathcal{S}$}
  \STATE Sample state $s_{sample} \sim P$
  \STATE $s_{near} \gets \texttt{nearest}(s_{sample},\mathcal{S},\hat c,1,\infty)$
  \STATE $s \gets \texttt{steer}(s_{sample}, s_{near}, \hat c, M)$
  \IF{$s \in \mathbb{S}_{free}$}
    \STATE $N \gets \texttt{nearest}(s,\mathcal{S},\hat c,\infty,M)$
    \STATE $\mathcal{E} \gets \mathcal{E} \cup \{(s,s') : s' \in N\}$
    \STATE $\mathcal{E} \gets \mathcal{E} \cup \{(s',s) : s' \in N\}$
    \STATE $\mathcal{V}(x) \gets \infty$
    \STATE $\mathcal{V} \gets \texttt{value\_iterate}(\mathcal{S},\mathcal{E},\mathcal{V},\hat c)$
  \ENDIF
\ENDWHILE
\RETURN $\mathcal{S},\mathcal{V}$
\end{algorithmic}}}
\caption{Function $\texttt{rrt\_sharp}(s_{start},s_{goal},p)$. The goal state $s_{goal}$ can be chosen arbitrarily from the goal region $S_{goal}$ of the original MDP. For notational simplicity, we omit the details of how the steering radius $M$ decreases as a function of the number of vertices $|\mathcal{S}|$.}
\label{alg:rrt}
\end{figure}

A few things are noteworthy about our implementation.
First, we search backwards from the goal to the start state.
This enables us to treat the search tree as a value function for MPC, since the goal state does not change during execution.

Second, rather than using D* to update the value function,
we simply perform brute-force sweeps of Bellman backups over the state set $\mathcal{S}$ using \texttt{value\_iterate}.
Practical implementations may wish to implement D* to speed up the planning time,
but the end result is the same.

Third, we perform collision checking only on the start and end points of each edge.
We assume that the maximum steering radius $M$ is small enough to ensure that no edges travel through thin obstacles.
The motivation for this is that we need to perform this operation with an even larger search radius $R > M$ during MPC, so it needs to be fast.

Fourth, we require $P(s_{start}) > 0$ in order to bias the samples towards the start state and to guarantee that the algorithm will terminate
(assuming a feasible path exists).
%Note that probabilistic completeness also requires that every open subset of $\mathbb{S}_{free}$ has nonzero probability.

\subsection{Extending the Horizon of MPPI }

\begin{figure}[t]
\centering
\framebox{\parbox{3in}{
\begin{algorithmic}
\REQUIRE Start state $s$
\REQUIRE Control sequence $a_{0:H-1}$
\REQUIRE Planning tree $\mathcal{S}$ with value function $\mathcal{V}$
\STATE Roll out $s_{0:H} = \hat f(s, a_{0:H-1})$
\STATE Find step costs $c_{step} = \sum_{t=0}^{H-1} c(s_t, a_t)$
\STATE Initialize terminal cost $c_{term} = \infty$
\STATE Find neighbors $N = \texttt{nearest}(s_H,\mathcal{S},\hat c,\infty,R)$
\IF {$N \neq \emptyset$}
  \STATE $c_{term} \gets \min_{s' \in N} \hat c(s_H,s') + \mathcal{V}(s')$
\ENDIF
\RETURN $c_{step} + c_{term}$
\end{algorithmic}}}
\caption{Function $\texttt{value}(s,a_{0:H-1},\mathcal{S},\mathcal{V})$.
For notational simplicity, we assume that collision detection is incorporated into the cost function $c(s,a)$.}
\label{alg:value}
\end{figure}

The main technical contribution of our approach is the \texttt{value} function shown in Fig.~\ref{alg:value},
which gives the exact process for interpreting the planning tree as a value function.
Given a starting state $s$ and control sequence $a_{0:H-1}$,
it essentially temporarily adds the terminal point of the rollout to the planning tree:
\begin{enumerate}
    \item Collect a list of the planning nodes within radius $R$ of the terminal point $s_H$, according to the distance function $\hat c$ used by the planner.
    \item For each of these nodes, calculate the exact cost of traveling to that node according to $\hat c$,\footnote{To save computation during MPC,
    we ignore obstacles when calculating the cost of traveling to a node.
    We assume that the search radius $R$ is small enough that the set of neighboring nodes does not include nodes on the other side of thin obstacles.}
    plus the value function $\mathcal{V}$ at that node according to the planner.
    \item Then the (approximate) value function $\widehat V(s_H)$ is given by the lowest of these summed costs.
\end{enumerate}

Note that if there are no planning nodes within radius $R$ of $s_H$, then the cost of that rollout is infinite.
Hence, the MPC controller must stay close to the planning tree $\mathcal{S}$  during execution,
else it risks becoming irrecoverably lost.

Fig.~\ref{alg:mppi} shows how the \texttt{value} function is used by MPPI.

\begin{figure}[t]
\centering
\framebox{\parbox{3in}{
\begin{algorithmic}
\REQUIRE Initial state $s_{start}$
\REQUIRE Goal region $S_{goal}$
\REQUIRE Planning tree $\mathcal{S}$ with value function $\mathcal{V}$
\REQUIRE Control covariance $\Sigma$
\REQUIRE Temperature $\lambda > 0$
\REQUIRE Number of trajectory samples $Q$
\STATE Initialize $\bar a_{0:H-1} = 0,\dots,0$
\STATE Initialize $s = s_{start}$
\WHILE{$s \notin S_{goal}$}
    \FOR{$i=1$ \TO $Q$}
       \STATE Sample $a^i_{0:H-1} \sim \mathcal{N}(\bar a_{0:H-1}, \Sigma)$
       \STATE $c^i \gets \texttt{value}(s, a^i_{0:H-1},\mathcal{S},\mathcal{V})$
       \STATE $w^i \gets \exp(-\frac{1}{\lambda} c^i)$
    \ENDFOR 
    \STATE $\bar a_{0:H-1} \gets \left( \sum_i w^i a^i_{0:H-1} \right) / \left( \sum_i w^i \right)$
    \STATE Execute $s \gets f(s,\bar a_0)$
    \STATE Shift $\bar a_{0:H-1} \gets \bar a_1, \bar a_2, \dots, \bar a_{H-1}, 0$ 
\ENDWHILE
\end{algorithmic}}}
\caption{Function $\texttt{mppi}(s_{start}, S_{goal}, \mathcal{S}, \mathcal{V}, \Sigma, \lambda, Q)$.
The distribution $\mathcal{N}(\mu,\Sigma)$ is a multivariate Gaussian with mean $\mu$ and covariance $\Sigma$.}\vspace{-4mm}
\label{alg:mppi}
\end{figure}

\section{EXPERIMENTS}
\label{sec:experiments}

\subsection{Kinematic 2D Environments}
\label{sec:kinematic}
We begin with several simple experiments using fully actuated planar robot models.
We consider both a simple point robot with state $s=(x,y)$ and a ``stick'' robot (shown in Fig. \ref{fig:dynamic} in blue) with an additional rotational degree of freedom, $\theta$.
Controls are state differences $a \in \mathbb{S}$ with cost
\begin{align}
\label{eq:mpc_cost}
c(s, a) = \begin{cases}
\mathbb{I}[s \notin S_{goal}] + (a^T W a)^{1/2}  & s \in \mathbb{S}_{free} \\
\infty & \text{otherwise}
\end{cases}
\end{align}
where $\mathbb{I}$ is a 0/1 indicator function, and $W \succ 0$ is a diagonal weight matrix.
We use MPPI for the MPC controller (cf. Fig. \ref{alg:mppi}) with the one-step dynamics model $\hat f(s,a) = s+a$.
The true dynamics $f$ are identical except that they implement noise by adding a small Gaussian perturbation to the control $a$.
During MPC execution, actions $a$ are clamped to have small norm $(a^T W a)^{1/2} < \varepsilon$ for some $\varepsilon > 0$.

The kinematic planning cost function is
\begin{align}
\label{eq:kinematic_cost}
\hat c(s_1,s_2) = \left( \matx{c}{x_2-x_1 \\ y_2-y_1 \\ \theta_2 - \theta_1}^T W \matx{c}{x_2-x_1 \\ y_2-y_1 \\ \theta_2 - \theta_1} \right)^{1/2}
\end{align}
which is similar to the MPC cost $c(s,a)$ except that it does not include the indicator term $\mathbb{I}[s \notin S_{goal}]$.
Instead, in the surrogate MDP that RRT\# is implicitly solving,
progress to the goal is encouraged by the fact that we \emph{must} take a step in the graph at every time step until we reach the goal.
Hence, the optimal policy is to take the shortest path to the goal.

We consider three different controllers:
\begin{itemize}
    \item \texttt{min}: MPPI with the subset of the planning tree $\mathcal{S}$ consisting of only the minimum-cost path
    \item \texttt{full}: MPPI with the full planning tree $\mathcal{S}$
    \item \texttt{naive}: A simple waypoint-based controller that simply aims straight at each subsequent node in the minimum-cost path.
\end{itemize}

\begin{figure}[t]
    \centering
    \includegraphics[width=0.23\textwidth]{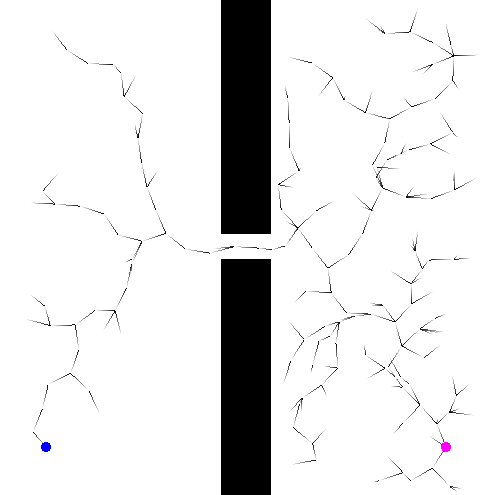}
    \includegraphics[width=0.23\textwidth]{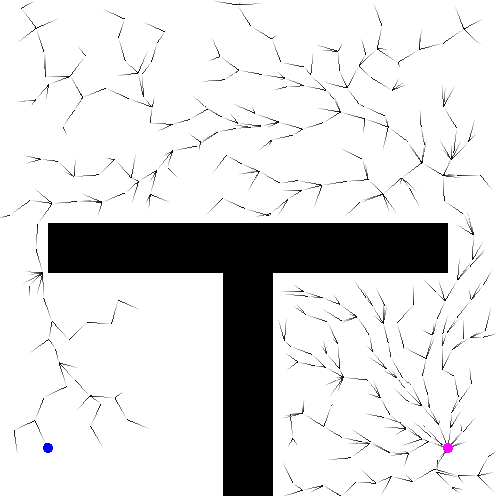}
    \includegraphics[width=0.23\textwidth]{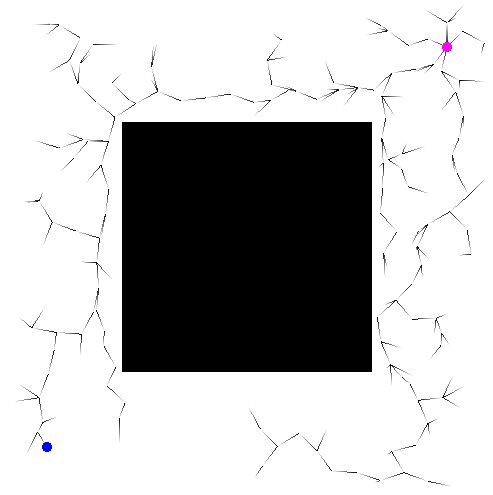}
    \includegraphics[width=0.23\textwidth]{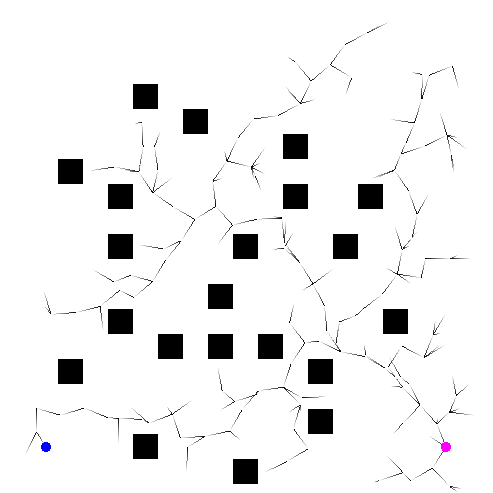}
    \caption{Planar planning environments. Clockwise from upper left: \texttt{gate}, \texttt{bugtrap}, \texttt{forest}, \texttt{blob}.
    The goal configuration is shown in magenta on the right side of each image.
    Also shown is the final RRT\# search tree for the point robot at the moment when a solution is found.}
    \label{fig:planar_tasks}
\end{figure}

For each of the four task environments shown in Fig. \ref{fig:planar_tasks}, we run the RRT\# planner 50 times,
to obtain a total of 200 planning trees.
For each planning tree, we run each controller for five trials.
We report three statistics from these trials:
\begin{itemize}
    \item \textbf{Failure \%}: We consider a trial to ``fail'' if the robot does not reach the goal within two minutes.
    This can happen either because the robot got stuck behind an obstacle,
    or because it drifted too far away from the planning tree and got lost.
    \item \textbf{Collision \%}: Of the successful trials, what percentage collided with an obstacle before reaching the goal?
    \item \textbf{Normalized Cost}: For a given planning tree, we calculate the normalized cost as follows.
    We compute the average cost accumulated by each controller over the five trials.
    (Failed or collided trials are not used in these averages.
    If a controller did not have at least three successful, collision-free trials, we exclude that planning tree from the the computation.)
    We divide each of these averages by the average for the \texttt{min} controller.
    Thus, the \texttt{min} controller always has normalized cost 1.0.
    We report the mean and standard deviation over the 200 planning trees.
\end{itemize}
Results are shown in Table \ref{table:static-results}.
As expected with randomized planning and control algorithms, there is a wide variance in normalized cost for the \texttt{full} controller.
However, overall it seems that in these very simple environments, the \texttt{full} controller slightly outperforms the \texttt{min} controller.
The \texttt{naive} controller collides with static obstacles on about one third of the trials, because it does not account for noise.
Interestingly, when it doesn't collide, it tends to find shorter paths than the MPC-based controllers.
Even though it cannot take shortcuts along the planned path like MPPI can,
it seems like the stochasticity of MPPI (or its ability to account for and avoid obstacles when the planned path goes near them) makes MPPI's paths longer.

Computationally, the \texttt{full} controller requires roughly 15 milliseconds (ms) of wall clock time per iteration. This is only about twice as expensive as the \texttt{min} controller (7 ms).
 
\begin{table}[ht]
\caption{Static experiment results}
\label{table:static-results}
\begin{center}
\begin{tabular}{|c||c|c|c|}
\hline
& \thead{Failure \%} & \thead{Collision \%} & \thead{Normalized Cost} \\
\hline
Point &
\makecell{\texttt{naive:} $0.0$\\ \texttt{min:~~} $0.0$\\ \texttt{full:~} $0.0$}
&
\makecell{\texttt{naive:} $34.6$\\ \texttt{min:~~} ~$0.0$\\ \texttt{full:~} ~$0.0$}
&
\makecell{\texttt{naive:} $0.950\pm0.023$\\ \texttt{min:~~} ---~~~~~~~~~~~~~~~\\ \texttt{full:~} $0.987\pm0.029$}
\\
\hline
Stick &
\makecell{\texttt{naive:} $0.0$\\ \texttt{min:~~} $2.2$\\ \texttt{full:~} $1.8$}
&
\makecell{\texttt{naive:} $34.0$\\ \texttt{min:~~} ~$1.0$\\ \texttt{full:~} ~$1.0$}
&
\makecell{\texttt{naive:} $0.917\pm0.090$\\ \texttt{min:~~} ---~~~~~~~~~~~~~~~\\ \texttt{full:~} $0.983\pm0.084$}
\\
\hline
\end{tabular}
\end{center}
\end{table}

\subsection{Dynamic 2D Environments}
\label{sec:second-order}

The usefulness of a value function becomes more obvious in environments where it is infeasible to stop and replan in the case of unforeseen circumstances.
To demonstrate this, we modify the previous experiments to incorporate second-order dynamics. We use state state $s = (p, \dot p)$ so that instead of the kinematic transition $f(s,a) = s+a$ we have
    \begin{align}
    \label{eq:second-order-dynamics}
        f\left(\matx{c}{p \\ \dot p}, a \right) = \matx{c}{p + \dot p \Delta t \\  \dot p + a}
    \end{align}
    where $\Delta t$ is the control timestep.
    After applying the update in Eq. (\ref{eq:second-order-dynamics}),
    the velocity $\dot p$ is clipped to have norm $(\dot p^T W \dot p)^{1/2} < \varepsilon$,
    so that we retain the same velocity limit as in the kinematic experiments.
    We again implement noise by perturbing the action $a$, and we use the same cost function $c(s,a)$ 
    with the same weights $W$ as in Eq. (\ref{eq:mpc_cost})
    except that the action $a$ is now in velocity space.
    The planner still uses the kinematic model and cost function $\hat c$ from Eq. (\ref{eq:kinematic_cost}),
    so in this case $c$ and $\hat c$ are only distantly related.
    
\begin{figure}[t]
    \centering
    \vspace{-6mm}
    \includegraphics[width=0.3\textwidth]{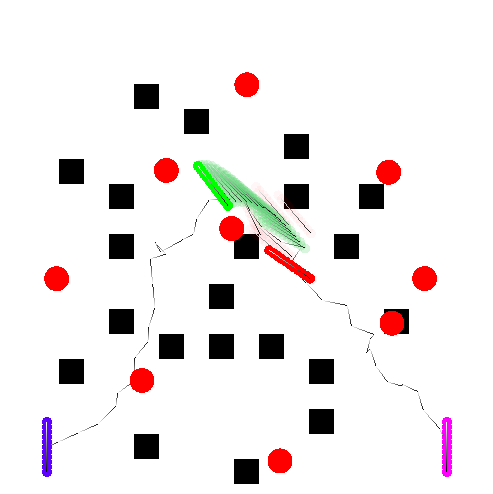}
    \caption{Visualization of MPC execution for the stick robot in the \texttt{forest} environment.
    The start and goal configurations are shown in blue and magenta, respectively.
    The thin, jagged line is a 2D projection (ignoring rotational pose) of the plan returned by the planner.
    Note that it is \emph{not} the full planning tree; rather, it is the shortest path from start to goal within that tree.
    The red circles are dynamic obstacles.
    The nominal MPPI trajectory is shown in in green.
    The planning node that gives the optimal terminal value for the nominal trajectory (cf. Fig. \ref{alg:value}) is shown in red.
    Finally, the very light lines near the end of the nominal trajectory are examples of terminal points $x_H$ from the most recent set of MPPI samples (cf. Figs. \ref{alg:value}, \ref{alg:mppi}).}\vspace{-4mm}
    \label{fig:dynamic}
\end{figure}

We further experiment with dynamic (i.e. moving / non-static) spherical obstacles which were not considered during planning,
shown in Fig. \ref{fig:dynamic} as red circles.
At each timestep, we perturb their velocity according to a uniform distribution
and then clamp it to a maximum speed.

Results are shown in Table \ref{table:dynamic-results}.
The results from Table \ref{table:static-results} are summarized in the row ``First / Static'' (meaning first-order dynamics / static obstacles).
Since the \texttt{naive} controller cannot handle dynamic obstacles or second-order dynamics, we report results only for \texttt{min} and \texttt{full}.
Interestingly, we see that both controllers are more successful and collide less frequently when the dynamics are second-order.
This may be because our implementation of noise in the second-order simulation is less aggressive,
or because MPC execution tends to take fewer time steps for the second-order system.
Perhaps the MPPI shift operator (cf. Fig. \ref{alg:mppi}) is more accurate for second-order systems.

The results with dynamic obstacles are very different.
For both first- and second-order systems, dynamic obstacles cause a dramatic increase in failure rate for the \texttt{min} controller,
but only a small increase in failure rate for the \texttt{full} controller.
Because we calculate normalized cost based on successful, collision-free trials,
the normalized cost of the \texttt{full} controller with dynamic obstacles varies considerably.
In fact, in the case of the second-order system, \texttt{full} finds slightly higher-cost trajectories than \texttt{min} on average.
Qualitatively, this is because during unlucky encounters with dynamic obstacles,
\texttt{full} is able to find (high-cost) alternative paths to avoid the dynamic obstacles,
while \texttt{min} simply collides or flies off of the planned path and cannot recover.
Failure can be interpreted as infinite cost, and is not included in normalized cost averages.

Qualitative results are \href{https://youtu.be/eB6i0V4Ij_0}{here} and source code is \href{https://github.com/nhatch/rrt/releases/tag/icra-2021}{here}.

\begin{table}[ht]
\caption{Dynamic experiment results}
\label{table:dynamic-results}
\begin{center}
\begin{tabular}{|c||c|c|c|}
\hline
& \thead{Failure \%} & \thead{Collision \%} & \thead{Normalized Cost} \\
\hline
\makecell{First / \\ Static} &
\makecell{\texttt{min:~~} $1.1$\\ \texttt{full:~} $0.9$}
&
\makecell{\texttt{min:~~} $0.5$\\ \texttt{full:~} $0.5$}
&
\makecell{\texttt{min:~~} ---~~~~~~~~~~~~~~~\\ \texttt{full:~} $0.985\pm0.063$}
\\
\hline
\makecell{First / \\ Dynamic} &
\makecell{\texttt{min:~~} $6.8$\\ \texttt{full:~} $1.4$}
&
\makecell{\texttt{min:~~} $2.0$\\ \texttt{full:~} $3.3$}
&
\makecell{\texttt{min:~~} ---~~~~~~~~~~~~~~~\\ \texttt{full:~} $0.947\pm0.273$}
\\
\hline
\makecell{Second / \\ Static} &
\makecell{\texttt{min:~~} $0.5$\\ \texttt{full:~} $0.5$}
&
\makecell{\texttt{min:~~} $0.0$\\ \texttt{full:~} $0.0$}
&
\makecell{\texttt{min:~~} ---~~~~~~~~~~~~~~~\\ \texttt{full:~} $0.973\pm0.049$}
\\
\hline
\makecell{Second / \\ Dynamic} &
\makecell{\texttt{min:~~} $3.0$\\ \texttt{full:~} $0.6$}
&
\makecell{\texttt{min:~~} $0.5$\\ \texttt{full:~} $0.4$}
&
\makecell{\texttt{min:~~} ---~~~~~~~~~~~~~~~\\ \texttt{full:~} $1.009\pm0.274$}
\\
\hline
\end{tabular}
\end{center}
\end{table}

\subsection{3D Skid Steering Environment}
\label{sec:warthog}

\begin{figure}[t]
    \centering
    \includegraphics[height=0.3\linewidth]{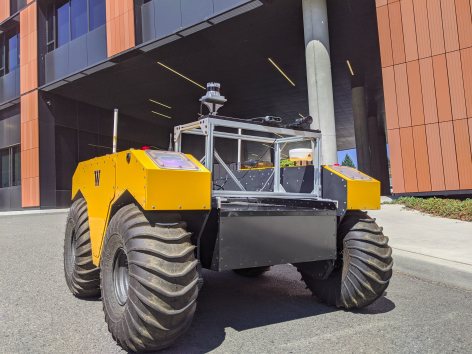}
    \includegraphics[height=0.3\linewidth]{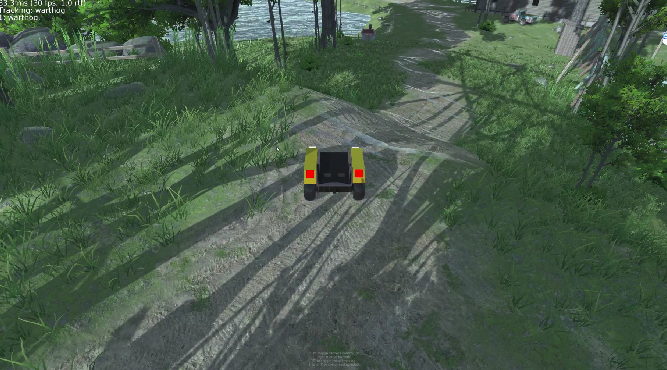}
    \caption{\textbf{Left:} The University of Washington Clearpath Warthog robot.
    \textbf{Right:} The 3D skid steering environment that simulates the Warthog.
    A video of this environment can be found 
 \href{https://youtu.be/a9_qFZR919k}{here}.
    The source code for this environment is not publicly available.}
    \label{fig:3d-env}
\end{figure}

Finally, we demonstrate the generality of this approach by showing a few examples in a 3D environment designed to simulate a real-world robot.
The Clearpath Warthog (see Fig \ref{fig:3d-env}, left) is a four-wheeled, skid-steered robot designed for off-road navigation.
The dynamics of this robot are difficult to model accurately, so we used a simple first-order kinematic model for both the RRT\# planner and the MPPI controller.
We implemented the algorithm proposed in this paper in the navigation stack for this robot, and we tested it in simulation (see Fig \ref{fig:3d-env}, right).

\begin{figure}
\centering
    \includegraphics[width=\linewidth]{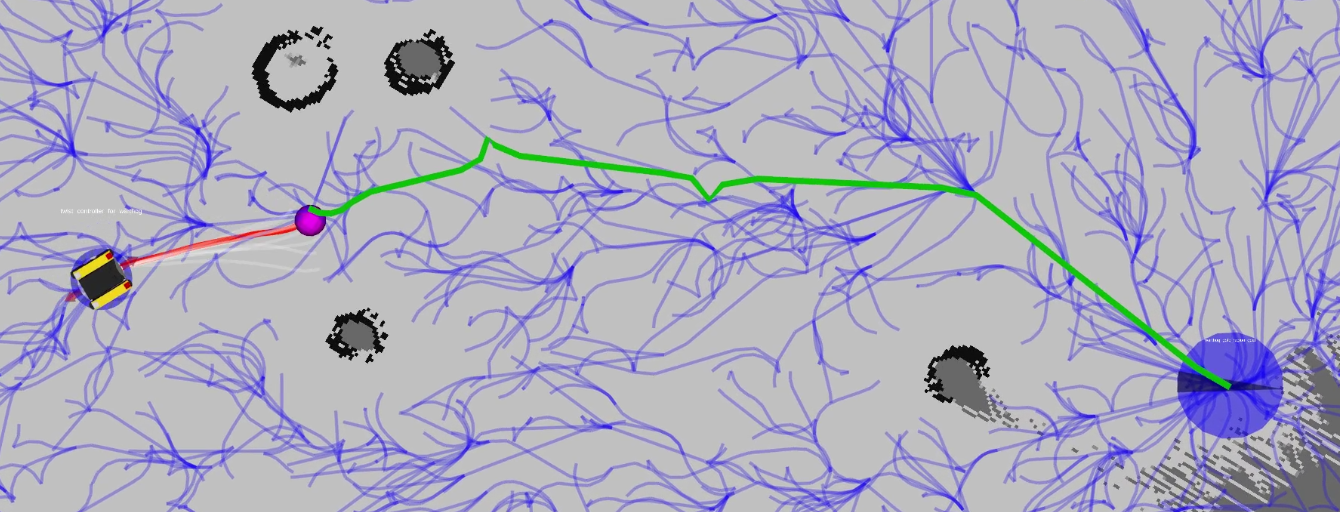} \\ ~\\
    \includegraphics[width=\linewidth]{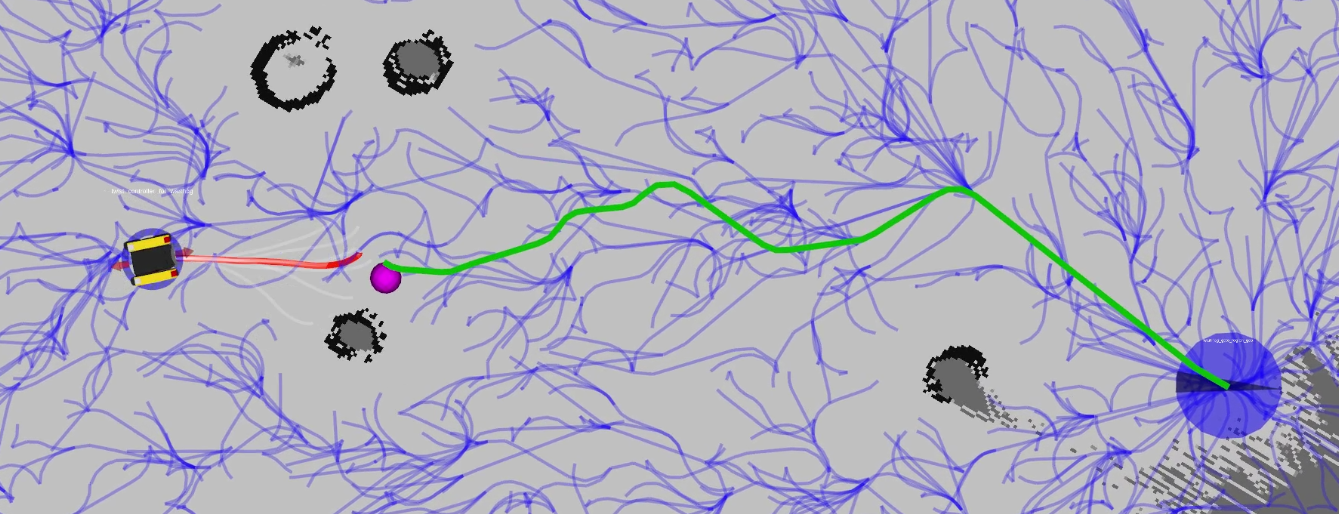}
    \caption{Qualitative results on the 3D environment.
    The optimized MPPI rollout is shown in red.
    The planning node used to compute the terminal cost for that rollout is shown in pink, and the corresponding path from that node to the goal is shown in green.
    The thin blue lines show other branches of the planning tree.
    See \href{https://youtu.be/xUW2hxHAh_A}{this video} for an animated version.}
    \label{fig:3d-qualitative}
\end{figure}

Fig. \ref{fig:3d-qualitative} shows qualitative results.
The lower screenshot was taken approximately one second after the upper one.
We see that, due to stochasticity and/or modeling errors, the robot's position has changed such that the optimized MPC rollout now uses a different branch of the planning tree.
Without access to the full planning tree, the MPC controller would not be able to make such optimizations.

\section{CONCLUSION}

In this paper, we have drawn a connection between reinforcement learning, control, and planning
by interpreting them all as searching for a value function.
The concept of a Bellman backup is central to all three,
although historically they have each used idiosyncratic names---e.g. dynamic programming, vertex expansion---for this concept.
This fundamental connection allows us to recognize \emph{planning trees} as approximate value functions
and reuse them as the terminal value function for MPC
at no additional computational cost.

\bibliographystyle{./bibliography/IEEEtran}
\bibliography{./bibliography/IEEEabrv,./bibliography/ref}

\end{document}